\pdfoutput=1
\documentclass[11pt]{article}

\usepackage[margin=1in]{geometry}
\usepackage[utf8]{inputenc}
\usepackage[T1]{fontenc}
\usepackage{lmodern}   
\usepackage{amsmath}
\usepackage{amssymb}
\usepackage{graphicx}
\usepackage{booktabs}
\usepackage{array}
\usepackage{tabularx}
\usepackage{float}     
\usepackage{multirow}
\usepackage{xcolor}
\usepackage{microtype}
\usepackage{authblk}
\usepackage[numbers,super,sort&compress]{natbib}
\makeatletter
\renewcommand{\@biblabel}[1]{#1.}
\makeatother
\usepackage[hidelinks]{hyperref}
\usepackage{xurl}      
\usepackage{titlesec}
\usepackage{enumitem}
\usepackage{caption}
\usepackage[section]{placeins}

\newlength{\figpanelwidth}
\newlength{\figtextwidth}
\newenvironment{figpanel}[1][0.56]
  {\global\setlength{\figpanelwidth}{#1\textwidth}%
   \global\setlength{\figtextwidth}{\dimexpr\textwidth-\figpanelwidth-0.03\textwidth\relax}%
   \noindent\begin{minipage}[t]{\figpanelwidth}\vspace{0pt}\centering}
  {\end{minipage}\hfill\ignorespaces}
\newenvironment{figtext}
  {\begin{minipage}[t]{\figtextwidth}\vspace{0pt}\setlength{\parskip}{0.4em}}
  {\end{minipage}}

\newcommand\samethanks[1][\value{footnote}]{\footnotemark[#1]}

\titleformat{\paragraph}[runin]{\normalfont\bfseries}{}{0em}{}[.]
\titlespacing*{\paragraph}{0pt}{1em}{0.6em}
\newcommand{\symgroup}[1]{\vspace{0.9em}\noindent\textbf{#1}\par\nobreak\vspace{0.35em}}

\title{\Large\bfseries Interpretable Symptom Vectors for Depression in a Large Language Model}

\author[1,2]{Fangyi Zhu\thanks{These authors contributed equally to this work.}}
\author[1,2]{Ajay Subramanian\samethanks}
\author[4]{Allison Constant}
\author[1,2]{Camille Wang}
\author[5]{Ravish Gupta}
\author[1,2,3]{Corey J. Keller}

\affil[1]{Department of Psychiatry \& Behavioral Sciences, Stanford University School of Medicine, Stanford, CA, USA}
\affil[2]{Wu Tsai Neurosciences Institute, Stanford University, Stanford, CA, USA}
\affil[3]{Veterans Affairs Palo Alto Healthcare System, and the Sierra Pacific Mental Illness, Research, Education, and Clinical Center (MIRECC), Palo Alto, CA, USA}
\affil[4]{Kaiser Permanente, San Jose, CA, USA}
\affil[5]{BigCommerce, Seattle, WA, USA}

\date{}

\begin{document}
\maketitle
\vspace{-2em}
\begin{center}
\textit{Corresponding author: Fangyi Zhu (\href{mailto:contact@fangyizhu.com}{contact@fangyizhu.com})}
\end{center}
\vspace{1em}

\begin{abstract}
Patients with depression present with diverse symptom profiles, yet clinical practice routinely reduces this variation to a single severity score. Large language models (LLMs) can potentially capture various symptoms and their severity from patient speech. However, how depressive symptoms are represented inside LLMs remains poorly understood, limiting clinical trust. To examine whether internal model activations match clinician judgment, we analyzed the residual stream of Gemma-3-27B-PT using mechanistic interpretability techniques. Recording activations across symptom descriptions drawn from validated clinical instruments, we found that symptom groups geometrically separated the most at layer 21 across multiple distance metrics. Using Semantic Projection, we then projected held-out naturalistic text onto Symptom Vectors constructed from these instruments. The resulting per-symptom coefficients preserved clinician-annotated rank ordering across mood, somatic, and suicidality axes. Furthermore, a single depression vector in Layer 21 separates held-out depressive from non-depressive text (AUC~$=$~0.789), which can be used as an emotional valence gate that restricts symptom projection to depressive speech. These results reveal a decorrelated, clinician-aligned symptom signal readable directly from internal activations, offering a mechanistic foundation for interpretable depression-assessment tools.
\end{abstract}

\section{Introduction}

Major depressive disorder spans highly diverse symptom profiles, biological underpinnings, and treatment trajectories, a multidimensional complexity that remains one of the most significant barriers to effective psychiatric care. In the STAR*D cohort, for example, 3{,}703 patients meeting the same DSM-5 criteria produced 1{,}030 unique symptom profiles\cite{Fried2015}. Standard questionnaires like PHQ and HAM-D prioritize brevity; they collapse rich symptom dimensions into a single score while overlooking symptoms that matter to patients\cite{Chevance2020}. These discarded nuances, however, are preserved when patients describe their own experiences in naturalistic language. Symptom-level extraction from such language could help differentiate patients with divergent underlying profiles and link linguistic markers to the neural and physiological biomarkers central to psychiatric neuroscience. Thus, symptom extraction from natural language has been a long-standing goal in clinical NLP.

Recent LLMs, which can take context into account and process text as a whole, have renewed interest in this goal. Existing systems typically produce broad diagnostic labels or sum scores rather than symptom-level assessments\cite{Yang2024,Xu2024,Lamichhane2023}. Generated outputs are also an unreliable window into model internals: GPT-4's internal schema of depression underemphasizes suicidality and overemphasizes psychomotor symptoms relative to expert and self-report ratings\cite{Ganesan2026}. Mechanistic interpretability techniques, however, offer a route to quantitative, per-symptom signals read directly from the model's internal representations. Prior work has shown that emotion representations are often encoded as approximately linear directions in an LLM's internal activations, with sentiment localized to specific layers in LLaMA\cite{Hollinsworth2024,DiPalma2025}. Moreover, the emotional information models pick up from text can be traced to specific, interpretable internal signals\cite{Tak2025}. To our knowledge, no prior work has mapped a clinically validated depression symptom taxonomy onto decorrelated per-symptom signals in an LLM's internal activations.

As an LLM processes text, it builds internal representations in a high-dimensional activation space known as the residual stream. Prior work suggests emotions are encoded as linear directions in this space; depression symptom categories may be similarly represented. Applying mechanistic interpretability techniques to Gemma-3-27B-PT, we show that these categories (mood/emotional/cognitive, somatic, and suicidality) are geometrically separable in the model's activations, with maximal separation appearing at layer 21. Projecting held-out naturalistic narratives onto clinician-derived Symptom Vectors yields coefficients that predict clinician category annotations. Furthermore, a single Depression Vector can distinguish between depressive and positive affect text. This study establishes a pipeline for extracting interpretable symptom signals directly from an LLM's internal state. These findings provide a methodological foundation for analyzing depression from patients' natural language, preserving the nuanced profiles that brief severity scores obscure. With longitudinal validation, such granular signals could support symptom tracking for research and, eventually, clinical use.

\section{Methods}

\begin{figure}[!htbp]
\centering
  \includegraphics[width=\textwidth]{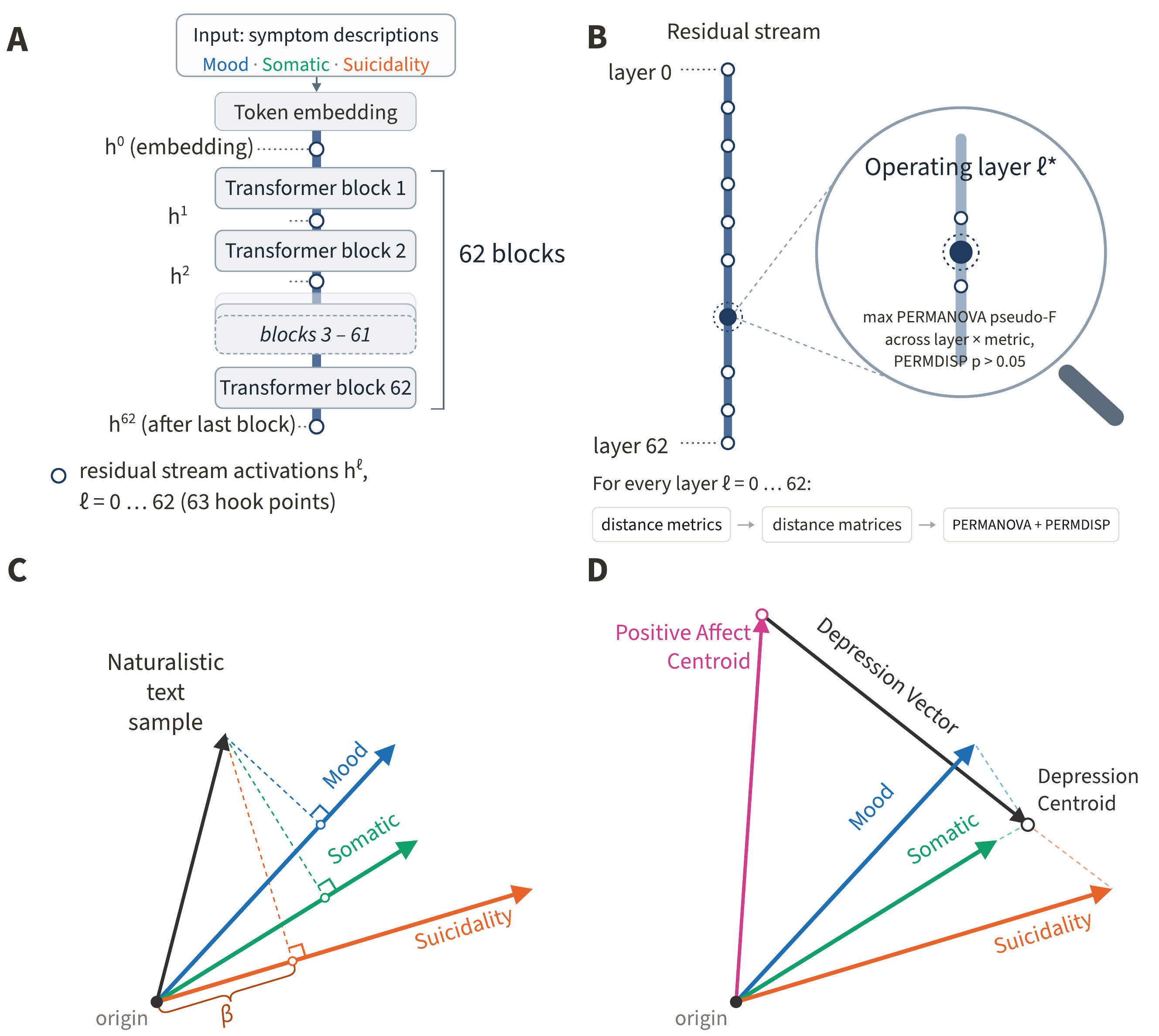}
  \captionsetup{width=\textwidth}
  \caption{\textbf{Analysis pipeline.} \textbf{A} Residual stream activations are extracted from Gemma-3-27B-PT across all 63 hook points ($\ell = 0,\dots,62$) while prompting with symptom descriptions from three groups. \textbf{B} For each layer $\times$ distance-metric combination, distance matrices are computed and tested with PERMANOVA and PERMDISP; the operating layer $\ell^*$ is the layer maximizing pseudo-$F$ subject to PERMDISP $P>0.05$. \textbf{C} Symptom Vectors are constructed from the group centroids at $\ell^*$, and naturalistic text samples are projected onto them to obtain symptom coefficients $\beta$. \textbf{D} The Depression Vector is defined as the difference between the Depression Centroid and the Positive Affect Centroid.}
  \label{fig:pipeline}
\end{figure}

As shown in Fig.~\ref{fig:pipeline}, our analysis consists of four stages: (1) extracting residual stream activations from Gemma-3-27B-PT for symptom descriptions, (2) selecting the operating layer where the three symptom groups are most separable, (3) constructing Symptom Vectors at that layer to project held-out text, and (4) contrasting the depression and positive affect centroids to derive a single Depression Vector.

\subsection{Data and preprocessing}

\subsubsection{Text Corpora}

The \textbf{Core Clinical} corpus was used to construct the Symptom Vectors. It comprised symptom descriptions extracted from six validated depression assessment instruments: the DSM-5\cite{APA2013}, ICD-10\cite{WHO1992}, Hamilton Depression Rating Scale (HAM-D)\cite{Hamilton1960}, Montgomery-\r{A}sberg Depression Rating Scale (MADRS)\cite{Montgomery1979}, Patient Health Questionnaire (PHQ-9)\cite{Kroenke2001}, and the patient-derived PROCEED outcome set\cite{Chevance2020}. Each instrument was segmented into shorter, self-contained excerpts. Each excerpt was assigned to a single symptom group by one psychiatrist author; a second psychiatrist independently reviewed all assignments against DSM-5 item definitions. Disagreements were resolved through joint review until the two annotators reached consensus, and a final label was assigned only by agreement. This yielded 51 excerpts (19 mood, 24 somatic, 8 suicidality) distributed across DSM-5 (3), ICD-10 (3), HAM-D (16), MADRS (12), PHQ-9 (9), and PROCEED (8).

We also assembled a \textbf{Naturalistic} corpus as held-out texts for Symptom Vector evaluation. This corpus comprised first-person and clinical-narrative descriptions of the same three symptom groups, drawn from three sources: William Styron's memoir \textit{Darkness Visible: A Memoir of Madness}\cite{Styron1990}, the \textit{Handbook of Depression}, 3rd Edition\cite{Gotlib2014}, and ReDSM5, a corpus of Reddit posts annotated at the sentence level against the nine DSM-5 depression symptoms by a licensed psychologist\cite{Bao2025}. We selected \textit{Darkness Visible} for its sustained first-person account of lived depressive experience and the \textit{Handbook of Depression} for its third-person clinical description of symptoms; ReDSM5 was used to supplement the books. \textit{Darkness Visible} and \textit{Handbook} excerpts were annotated using the same two-author categorize-then-review procedure for the Core Clinical corpus. ReDSM5 samples, inheriting DSM-5 labels assigned by a licensed psychologist\cite{Bao2025}, were programmatically mapped onto the three symptom groups under the same DSM-5-item assignment. To balance the symptom groups, we retained every book-sourced passage and added randomly selected ReDSM5 posts to each group until its count matched that of the largest book-sourced group. Table~\ref{tab:naturalistic} reports the resulting sample counts.

\begin{table}[!htbp]
\centering
\caption{Naturalistic corpus composition. Excerpt counts by source and symptom group (mood, somatic, suicidality).}
\label{tab:naturalistic}
\begin{tabular}{lrrrr}
\toprule
Source & Mood & Somatic & Suicidality & Source Total \\
\midrule
Darkness Visible        & 33  & 11  & 13  & 57  \\
Handbook of Depression  & 108 & 42  & 22  & 172 \\
ReDSM5                   & --- & 88  & 106 & 194 \\
\midrule
Symptom Total            & 141 & 141 & 141 & 423 \\
\bottomrule
\end{tabular}
\end{table}

Excerpts of the Core Clinical and Naturalistic corpora were annotated with one of three symptom groups: mood/emotional/cognitive (hereafter ``mood''; DSM-5 A1, A2, A7, A8: depressed mood, anhedonia, guilt or shame, concentration difficulties); somatic (DSM-5 A3--A6: appetite or weight change, sleep disturbance, psychomotor change, and fatigue in the context of depression); and suicidality (DSM-5 A9: active suicidal ideation with or without specific plan, and passive death wishes). The mood and somatic groups were chosen to reflect depression's distinct impact on the mind or body, following prior meta-analytic work on clinical instruments\cite{Shafer2006}. Given its clinical significance and risk profile, suicidality was assigned its own category.

The \textbf{Positive Affect} corpus comprises short excerpts of a few sentences expressing positive affect such as joy, love, wonder, aspiration, and optimism, from deliberately diverse sources spanning translated classical Chinese poetry (Li Bai, Cao Cao), literary fiction (Tolstoy, Garc\'ia M\'arquez, Rothfuss), popular-science writing (Dawkins), political oratory (Kennedy), and foundational civic prose (the U.S.\ Constitution). This corpus was constructed to be opposite in valence relative to the depression corpora: it consists of emotionally positive text, whereas the depression corpora are negative. It also spans a broad range of genres and styles, so that any separation observed along the symptom axes reflects shared affective valence. There are 9 total excerpts in this group. The plain text of this corpus is included in the GitHub repository.

To contrast depressive text with affectively positive language, we included \textbf{HappyDB}, a corpus of crowdsourced descriptions of happy moments\cite{Asai2018}. We took the first 423 non-duplicate entries to match the size of the Naturalistic corpus.

\subsubsection{Text Anonymization and Standardization}

A pilot experiment on the unedited Core Clinical corpus showed that PROCEED’s residual stream is geometrically separated from all other clinical instruments regardless of clinical similarity. This separation is an artifact of its distinctive format of bullet-point symptom lists. To prevent stylistic confounds such as format, pronouns, and register from being picked up as signal, we rewrote every excerpt into a standard third-person register using Anthropic Claude Opus 4. All excerpts were rewritten with the same prompt: ``Rewrite the following text into natural language as a description of a third person. Correct any grammar mistakes.'' All rewrites were reviewed and corrected by the authors.

These rewrites were deliberately minimal, comprising surface-level edits only: standardizing pronouns, removing severity ratings from clinical questionnaires, correcting grammar in ReDSM5 and HappyDB, and restructuring fragments into continuous prose in PROCEED. Names and identifying details were also removed. Clinical content was left unchanged.

The full set of rewritten excerpts in the Core Clinical and Positive Affect corpora is provided in the supplementary materials. Table~\ref{tab:standardization} shows representative examples of the rewriting from all corpora.

\begin{table}[H]
\centering
\caption{Text standardization examples. Original and standardized excerpts from each publicly available corpus.}
\label{tab:standardization}
\footnotesize
\setlength{\extrarowheight}{1pt}
\begin{tabularx}{\linewidth}{l X X}
\toprule
Dataset & Original & Standardized \\
\midrule
Core Clinical &
INSOMNIA -- Delayed (Waking in early hours of the morning and unable to fall asleep again) \newline 0 = Absent \newline 1 = Occasional \newline 2 = Frequent &
Delayed Insomnia: The person wakes up in the early hours of the morning and is unable to fall asleep again. \\
\addlinespace[10pt]
\begin{tabular}[t]{@{}l@{}}Naturalistic\\(Darkness Visible)\end{tabular} &
I now see, as a means to calm the anxiety and incipient dread that I had hidden away for so long somewhere in the dungeons of my spirit. &
They now see this as a means to calm the anxiety and incipient dread that they had hidden away for so long somewhere in the dungeons of their spirit. \\
\addlinespace[10pt]
\begin{tabular}[t]{@{}l@{}}Naturalistic\\(Handbook of Depression)\end{tabular} &
Individuals with depression exhibit cognitive deficits and biased processing of emotional material. &
They exhibit cognitive deficits and biased processing of emotional material. \\
\addlinespace[10pt]
Positive Affect &
He spoke gently, laughed often, and never exercised his wit at the expense of others. &
They speak gently, laugh often, and never exercise their wit at the expense of others. \\
\addlinespace[10pt]
HappyDB &
I went on a successful date with someone I felt sympathy and connection with. &
They went on a successful date with someone they felt sympathy and a connection with. \\
\bottomrule
\end{tabularx}
\end{table}
\subsubsection{Residual Stream Extraction}

For this study, we used Google's official Hugging Face release of Gemma-3-27B-PT\cite{Gemma2025}, a 27.4-billion-parameter model with 62 transformer blocks and a residual stream width of 5{,}376. We used the pre-trained model rather than the instruction-tuned variant, since post-training safety alignment could distort the model's representation of depression symptoms.

Each excerpt was tokenized with the Gemma-3 SentencePiece tokenizer and processed in a single forward pass through the model, which was loaded and run with the Hugging Face Transformers library\cite{Wolf2020} on a PyTorch backend\cite{Paszke2019} using Apple's Metal Performance Shaders (MPS). During each pass, forward hooks recorded the input embeddings and the residual stream output of every transformer block, yielding 63 activations in total. Activations corresponding to the initial \texttt{<bos>} token were discarded at every hook point prior to all subsequent analyses, to mitigate the attention-sink and massive-activation phenomena documented in transformer residual streams\cite{Xiao2024,Sun2024}.

\subsection{Per Residual Stream Layer Separability}

Different residual stream layers encode different information. To identify where the model distinguishes depression symptoms, we quantified symptom-group separability at each residual stream layer on the Core Clinical corpus using PERMANOVA. We then used the layer with the highest pseudo-$F$ for the Symptom Vector extraction.

\subsubsection{Pairwise Distance Matrices}

Residual stream activations are high-dimensional (5{,}376) objects of varying length, whose token activations depend on preceding tokens through the self-attention mechanism. Tokens within an activation are therefore not independent samples; each activation has to be compared as a whole. There is no established method for measuring the distance between residual streams generated by prompts of different lengths for the purpose of comparing textual meaning. We tested the efficacy of eight distance-metric $\times$ normalization combinations: Centroid Cosine Distance, Centroid Euclidean Distance, Earth Mover's Distance, and Energy Distance, each computed on both raw and $\ell^2$-normalized residual stream activations, at every residual stream layer $\ell \in \{0,1,\dots,62\}$.

These metrics are not directly comparable, so we reduced each to a common representation: a pairwise distance matrix. For a given metric and layer, entry $(i,j)$ is the distance between the residual stream activations of prompts $i$ and $j$ from the Core Clinical corpus. The distance matrices rule out classical ANOVA and MANOVA, which need fixed-length coordinate vectors and normally distributed underlying data, neither of which residual streams satisfy. PERMANOVA operates directly on the pairwise distances, and its pseudo-$F$ statistic is sensitive to the distance between group centroids. Because pseudo-$F$ is a ratio of between- to within-group variation, it is unaffected by the absolute scale of the input distances. PERMANOVA can thus be used to compare different metrics directly.

The centroid metrics use the mean of all token activations. Let $h_t^\ell$ denote the residual stream activation of token $t$ at layer $\ell$, for a sequence of $T$ tokens. The layer's centroid is defined as:
\begin{equation}
\bar{h}^\ell = \frac{1}{T}\sum_{t=1}^{T} h_t^\ell.
\label{eq:centroid}
\end{equation}
The centroid distance is then computed using PyTorch tensor operations\cite{Ansel2024}.

Distribution-based metrics (Earth Mover's, Energy) were computed directly between full token distributions. Energy Distance was computed with the \texttt{pted} package\cite{Stone2026}, an implementation of the energy statistic with permutation tests\cite{Szekely2013}. Earth Mover's distance was computed with SciPy\cite{Virtanen2020}.

\subsubsection{PERMANOVA with the PERMDISP eligibility gate}

Using the pairwise distance matrices, we assessed the separability of residual stream activations across symptom groups with PERMANOVA, gated by PERMDISP. At each layer and for each metric, we computed a PERMANOVA pseudo-$F$ statistic\cite{Anderson2001}:
\begin{equation}
\text{pseudo-}F = \frac{\sum(\text{Between-Group Distance})^2 / (g-1)}{\sum(\text{Within-Group Distance})^2 / (n-g)},
\label{eq:permanova}
\end{equation}
where $g=3$ is the number of symptom groups and $n$ is the total number of excerpts.

PERMANOVA is sensitive not only to differences in group centroid location but also to differences in within-group multivariate dispersion: a group whose excerpts are widely spread can produce a significant pseudo-$F$ even when its centroid coincides with the others. A pseudo-$F$ is therefore interpretable as centroid separation only at layers where within-group dispersion is homogeneous across groups. To identify those layers, we computed the Permutational Analysis of Multivariate Dispersions (PERMDISP)\cite{Anderson2006}, which tests the null hypothesis of equal within-group dispersion, at every layer. We define the \emph{eligible region} of a metric as the set of layers at which PERMDISP is non-significant ($P>0.05$), meaning there is no significant difference in group dispersion; at these layers a PERMANOVA pseudo-$F$ reflects genuine displacement of group centroids. All distance statistics were computed with the scikit-bio package\cite{Aton2026} (PERMDISP with \texttt{test="centroid"}), each on 9{,}999 permutations.

\subsubsection{Operating Layer selection}

For each metric, the operating layer is the layer of highest PERMANOVA pseudo-$F$ within the eligible region. All eight metric--normalization combinations follow this identical procedure and are ordered by their operating PERMANOVA pseudo-$F$. At each operating layer we report the PERMANOVA pseudo-$F$ and $P$-value and the PERMDISP $F$ and $P$-value; a PERMDISP $P$ just above 0.05 marks weakly supported eligibility.

\subsection{Semantic Projection for Symptom Identification}

The procedure below adapts the semantic projection framework of Grand et al.\cite{Grand2022}, who showed that context-dependent human judgments about object features can be recovered from word embeddings by projecting word vectors onto interpretable axes: the line connecting ``small'' to ``big,'' for instance, yields an axis along which a word's projection reflects perceived size. We extend this framework in three ways. First, symptom axes are constructed from class-centroid activations rather than antonymous word pairs, because clinical symptom categories (mood, somatic, suicidality) lack the natural antonymy of adjective pairs. Second, the three symptom axes are projected jointly rather than independently, with the resulting values decorrelated via a Gram-pseudoinverse correction. Third, the representations being projected are residual stream activations of a contemporary transformer language model, taken at the operating layer and in the raw activation geometry identified in the previous experiment, whereas the original framework used static distributional word vectors.

The output of the procedure is, for each input passage $x$, a three-dimensional vector of decorrelated projection coefficients,
\begin{equation}
\beta(x) = (\beta_{\text{mood}}, \beta_{\text{somatic}}, \beta_{\text{suicidality}}) \in \mathbb{R}^3,
\label{eq:beta-def}
\end{equation}
each coefficient quantifying the unique contribution of one symptom axis to the passage's representation in residual stream space at the operating layer. We refer to the subspace of residual stream space spanned by the three symptom basis vectors as the \emph{symptom subspace}. The construction of the basis, the projection, and the properties of $\beta$ are described in turn below.

\paragraph{Symptom Vector Construction} At the operating layer $\ell^*$ identified by the separability analysis, we constructed three Symptom Vectors $v_s \in \mathbb{R}^{5376}$ for $s \in \{\text{mood}, \text{somatic}, \text{suicidality}\}$ from the Core Clinical corpus. Each $v_s$ is the centroid of all residual stream activations associated with symptom $s$, computed in the model's native activation geometry. As throughout, the initial \texttt{<bos>} token is removed before averaging, to prevent attention-sink effects\cite{Xiao2024,Sun2024}, and all centroid arithmetic is performed in float64 on CPU. The three centroids were stacked to form the basis matrix:
\begin{equation}
V = [\,v_{\text{mood}};\, v_{\text{somatic}};\, v_{\text{suicidality}}\,] \in \mathbb{R}^{3\times 5376}.
\label{eq:basis}
\end{equation}

\paragraph{Gram matrix construction} Because the axes are nearly collinear, a naive projection conflates their contributions; we therefore decorrelate the projection using the inverse of the Gram matrix. The geometric relationships between $v_{\text{mood}}, v_{\text{somatic}}, v_{\text{suicidality}}$ are captured by the Gram matrix, formed from the basis matrix $V$ constructed in the previous step:
\begin{equation}
G = V V^{\top} \in \mathbb{R}^{3\times 3}.
\label{eq:gram}
\end{equation}
Its entries are the pairwise inner products $G_{ij} = v_i \cdot v_j$, so the diagonal $G_{ii} = \lVert v_i \rVert^2$ holds the squared norms of the Symptom Vectors and the off-diagonal entries hold their overlaps. An orthonormal basis would give $G=I$. As the basis approaches parallel, the smallest eigenvalue of $G$ approaches zero and the condition number $\kappa(G) = \lambda_{\max}/\lambda_{\min}$ becomes large, and inverting $G$ amplifies small perturbations in the embeddings into large errors in the symptom loadings. This near-singular condition is addressed by the pseudoinverse below.

\paragraph{Pseudoinverse of the Gram matrix} Using $G^{-1}$ is not practical here. The clinical symptom categories of depression are textually comorbid: a passage describing one symptom often touches on others, so the class-centroid basis is expected to be near-collinear. $G$ is then ill-conditioned, and $G^{-1}$ amplifies small perturbations in the projected coordinates by a factor of up to $\kappa(G)$. If the basis vectors were exactly parallel, $G$ would be singular and $G^{-1}$ would not exist at all.

We therefore use the Moore--Penrose pseudoinverse $G^{+}$ in place of $G^{-1}$. For invertible $G$, $G^{+}=G^{-1}$; for singular or near-singular $G$, $G^{+}$ yields the unique $\beta$ minimizing both the residual $\lVert V c(x) - G\beta \rVert^2$ and the coefficient norm $\lVert \beta \rVert^2$. The condition number of $G$ in this study is reported in \S\ref{sec:results}.

\paragraph{Symptom Projection} Finally, we reduce each excerpt from a residual stream of $5{,}376$ dimensions $\times$ excerpt length to 3 symptom coefficients. The decorrelated projection coefficients for a passage $x$ are:
\begin{equation}
\beta(x) = G^{+} V\, c(x),
\label{eq:projection}
\end{equation}
where $c(x)$ is the passage's centroid representation: we extracted the operating-layer residual stream activations for $x$ and computed the float64 centroid of the remaining tokens, using the identical procedure as for the symptom basis vectors (Symptom Vector Construction, above).

This formulation is the multi-axis analogue of the projection step in Grand et al.\cite{Grand2022}: when the basis is orthonormal, $G^{+}=I$ and $\beta$ reduces to an inner-product projection onto each axis independently. When the basis is collinear, as is generally the case for class-centroid bases over correlated taxonomies, the pseudoinverse generalizes that projection, recovering each axis's unique contribution after accounting for the others.

The resulting projection coefficients $\beta$ have three interpretable properties:

\begin{itemize}[leftmargin=1.4em]
\item \textbf{Per-symptom isolation.} Each $\beta_s$ reflects the unique contribution of $v_s$ alone, with the shared signal among correlated axes apportioned by the Gram correction.
\item \textbf{Sign.} The sign of $\beta_s$ indicates whether the passage points along $v_s$ (positive) or against it (negative).
\item \textbf{Cross-passage comparability.} Because $\beta$ is derived from a fixed basis $V$ and a uniform projection procedure, projection coefficients from different excerpts and corpora live in the same coordinate system, enabling direct cross-corpus comparison, e.g., comparing $\beta_{\text{suicidality}}$ between clinical instrument and naturalistic depression descriptions.
\end{itemize}

We projected each passage's centroid from the Core Clinical and Naturalistic corpora onto $V$ using this procedure. Core Clinical projections constitute an in-sample consistency check; only the Naturalistic projections evaluate whether the methodology generalizes to held-out text as external validation. Each passage thus received three decorrelated projection coefficients $(\beta_{\text{mood}}, \beta_{\text{somatic}}, \beta_{\text{suicidality}})$, interpreted in \S\ref{sec:results}.

\subsection{Depressive and Positive Affect Distinction}

Because projecting non-depressive text onto Symptom Vectors is not meaningful, we also tested whether an LLM can distinguish between depressive and non-depressive text. We performed a control experiment at the operating layer. Two centroids of opposite emotional valence were built from the operating layer:

\begin{enumerate}[leftmargin=1.6em]
\item The \textbf{Depression} centroid is the centroid of the three Symptom Vectors from the previous experiment, constructed from the Core Clinical corpus. This gives equal weight to all three symptoms:
\begin{equation}
\bar{h}_{\text{depressive}} = \frac{v_{\text{mood}} + v_{\text{somatic}} + v_{\text{suicidality}}}{3}.
\label{eq:depcentroid}
\end{equation}
\item The \textbf{Positive Affect} centroid is the token-weighted mean over all tokens of the entire Positive Affect corpus, excluding \texttt{<bos>}:
\begin{equation}
\bar{h}_{\text{positive}} = \frac{1}{N_{P}} \sum_{x \in P} \sum_{t=1}^{T_x} h_{x,t}, \qquad N_{P} = \sum_{x \in P} T_x,
\label{eq:posaffect}
\end{equation}
where $P$ is the set of positive-affect texts, $T_x$ is the number of tokens in text $x$, $t$ indexes the tokens within a text, and $N_{P}$ is the total token count across the set.
\end{enumerate}

We then define the \textbf{Depression Vector} to be:
\begin{equation}
v_{\text{depression}} = \bar{h}_{\text{depressive}} - \bar{h}_{\text{positive}}.
\label{eq:depvector}
\end{equation}
The Naturalistic corpus and the HappyDB corpus are reserved as held-out tests that aren't included in the construction of these axes.

An excerpt $x$ is scored by the cosine similarity between its token-weighted centroid $\bar{h}_x$ (computed as above, excluding \texttt{<bos>}) and the Depression Vector; we call this the \textbf{Depression Score} of excerpt $x$:
\begin{equation}
\text{Score}(x) = \cos\!\big(\bar{h}_x, v_{\text{depression}}\big) = \frac{\bar{h}_x \cdot v_{\text{depression}}}{\lVert \bar{h}_x \rVert \, \lVert v_{\text{depression}} \rVert}.
\label{eq:score}
\end{equation}

We compared Depression Scores between the held-out Naturalistic and HappyDB corpora (423 excerpts each) using a one-sided Mann--Whitney $U$ test. We report the test statistic in its normalized form $U/(n_1 n_2)$, where $n_1$ and $n_2$ are the number of excerpts in the two corpora being compared, which equals the area under the receiver operating characteristic curve (AUC) and is interpretable as a probability of superiority: the chance that a randomly drawn depressive excerpt scores higher than a randomly drawn non-depressive one. A value of 0.5 indicates no separation; 1.0 indicates complete separation. We computed the same statistic for Core Clinical versus Positive Affect as an in-sample consistency check.

\section{Results}
\label{sec:results}

\subsection{Symptom categories are most separable at layer 21}

\begin{figure}[H]
\centering
  \includegraphics[width=\textwidth]{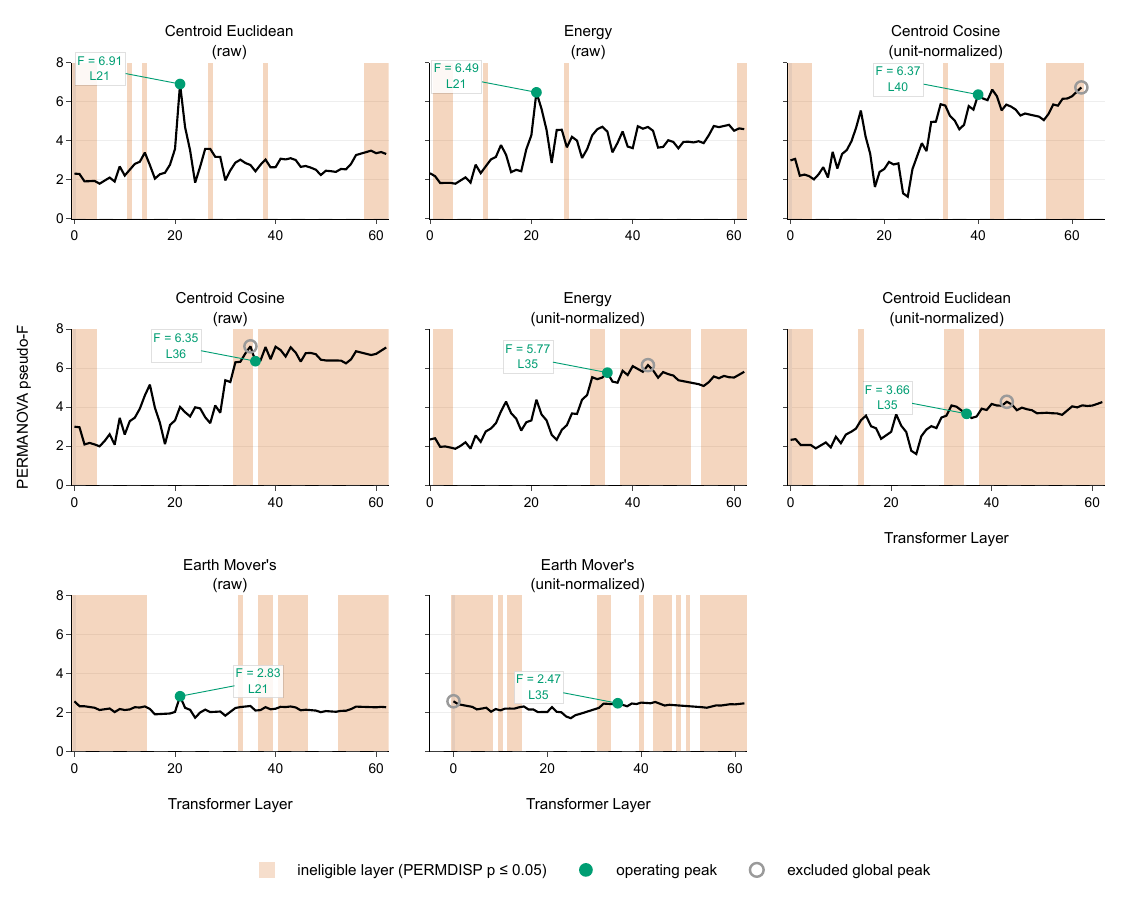}
  \captionsetup{width=\textwidth}
  \caption{\textbf{Layer 21 maximizes symptom separability.} PERMANOVA pseudo-$F$ ($y$-axis) across all 63 hook points ($x$-axis) for each of the eight distance-metric $\times$ normalization combinations. Shaded regions mark layers where PERMDISP is significant ($P<0.05$) and pseudo-$F$ is therefore ineligible as a measure of centroid separation; the operating maximum (filled marker) is the maximum pseudo-$F$ within each metric's eligible region ($P>0.05$), and the excluded maximum (open marker) is a higher but ineligible global maximum where shown.}
  \label{fig:separation}
\end{figure}

We first asked whether depression-symptom categories are geometrically separable in the model's residual stream and, if so, at which layer that separability is strongest. Across all 63 hook points, we computed the PERMANOVA pseudo-$F$ and its corresponding PERMDISP $P$-value for the eight distance-metric $\times$ normalization combinations. For each combination, we then selected the maximum pseudo-$F$ within the eligible region (Fig.~\ref{fig:separation}; Table~\ref{tab:permanova}).

\begin{table}[H]
\centering
\caption{PERMANOVA and PERMDISP statistics at each metric's eligible maximum. For each distance-metric $\times$ normalization combination, we tabulated the residual stream layer where PERMANOVA pseudo-$F$ is maximized within the eligible region (PERMDISP $P>0.05$).}
\label{tab:permanova}
\small
\begin{tabular*}{\textwidth}{@{\extracolsep{\fill}}llrrrrr@{}}
\toprule
 & & & \multicolumn{2}{c}{PERMANOVA} & \multicolumn{2}{c}{PERMDISP} \\
\cmidrule(lr){4-5}\cmidrule(lr){6-7}
Metric & Normalization & Layer & pseudo-$F$ & $P$ & $F$ & $P$ \\
\midrule
\multirow{2}{*}{Centroid Cosine}    & raw            & 36 & 6.35 & $<0.0001$ & 3.66 & 0.0686 \\
                                     & $\ell^2$-normalized & 40 & 6.37 & $<0.0001$ & 3.76 & 0.0832 \\
\multirow{2}{*}{Centroid Euclidean} & raw            & 21 & 6.91 & 0.0002    & 1.24 & 0.4313 \\
                                     & $\ell^2$-normalized & 35 & 3.66 & 0.0005    & 3.83 & 0.1104 \\
\multirow{2}{*}{Earth Mover}        & raw            & 21 & 2.83 & $<0.0001$ & 1.20 & 0.4713 \\
                                     & $\ell^2$-normalized & 35 & 2.47 & 0.0016    & 2.92 & 0.1925 \\
\multirow{2}{*}{Energy}             & raw            & 21 & 6.49 & 0.0010    & 0.68 & 0.6488 \\
                                     & $\ell^2$-normalized & 35 & 5.77 & $<0.0001$ & 4.58 & 0.0566 \\
\bottomrule
\end{tabular*}
\end{table}

Three observations identify layer 21 as the operating layer for the subsequent experiments. First, the three raw distance metrics converged on the same layer: Centroid Euclidean, Energy, and Earth Mover's distance each reached their eligible pseudo-$F$ maximum at layer 21. Second, at layer 21 the PERMDISP $P$-values for all three distance metrics were far from significance ($P=0.43$, $0.65$, and $0.47$), so the pseudo-$F$ at this layer is interpretable as centroid displacement rather than unequal dispersion. Third, the largest single eligible pseudo-$F$ across the entire sweep, 6.91 for Centroid Euclidean (raw), occurred at layer 21. We therefore adopt layer 21 as the operating layer $\ell^*$ for symptom-vector construction and projection.

The remaining metrics' separation maximized late in the network but with smaller centroid differences. The cosine metrics maximized at layers 36--40, and the $\ell^2$-normalized Euclidean, Energy, and Earth Mover's variants maximized at layer 35. At each of these later maxima, PERMDISP sat narrowly above the eligibility threshold ($P>0.05$), so part of the separation at these layers is due to different group dispersion. The three raw metrics agree at layer 21, making it a more reliable operating point than the weakly eligible maxima near layer 35.

\clearpage

\subsection{Semantic Projection coefficients align with clinician ratings}

\begin{figure}[!htbp]
\centering
  \includegraphics[width=\textwidth,trim=0 20 0 12,clip]{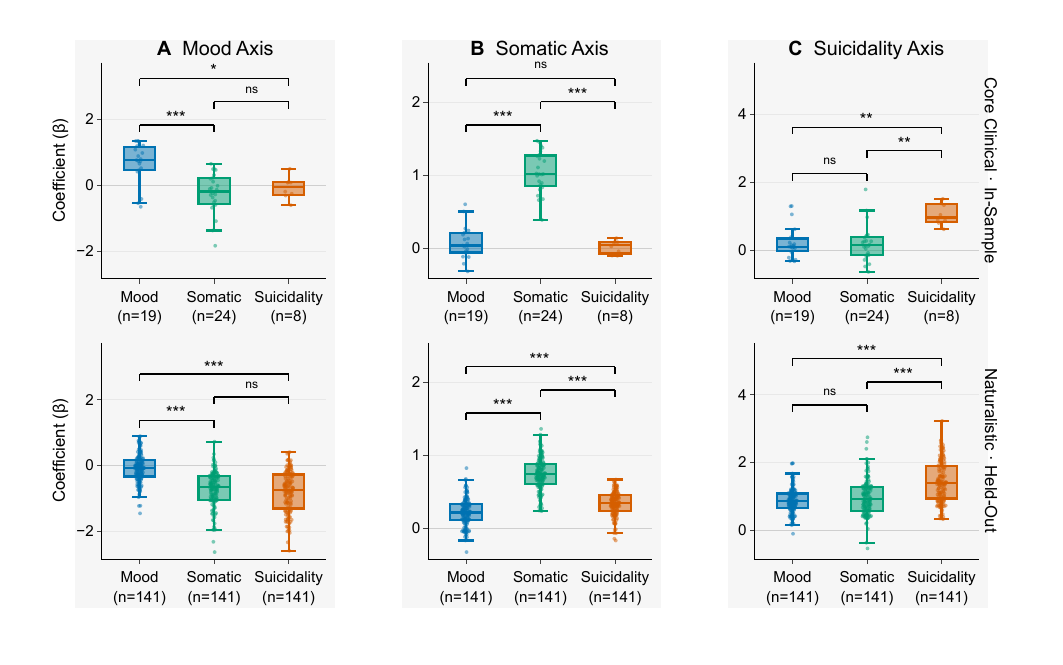}\\[0.4em]
  \includegraphics[width=\textwidth]{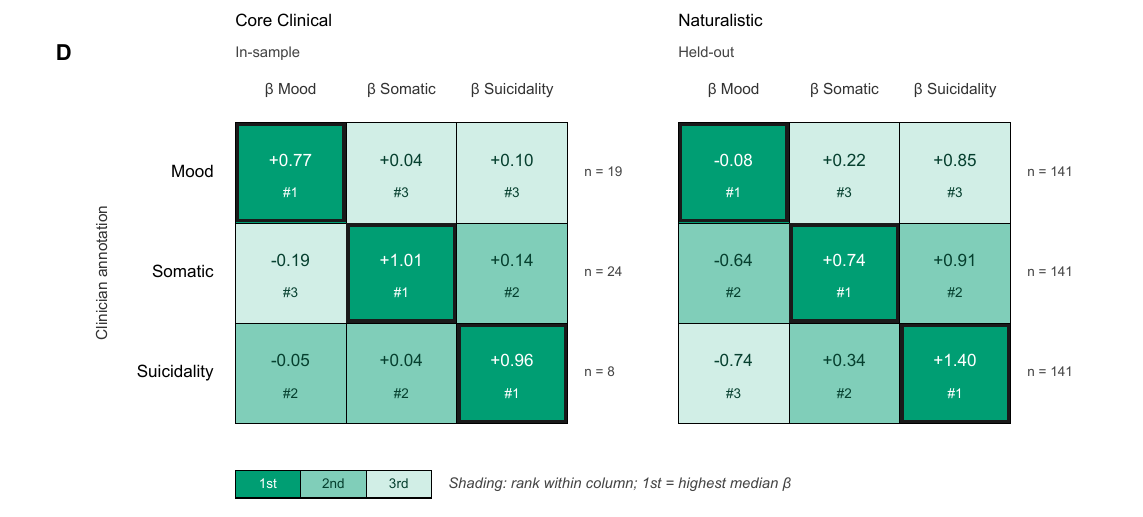}
\vspace{0.6em}
\caption{\textbf{Semantic projection yields clinician-aligned symptom coefficients.} Symptom coefficients $(\beta_{\text{mood}}, \beta_{\text{somatic}}, \beta_{\text{suicidality}})$ projected onto Symptom Vectors, grouped by clinician annotation and decorrelated. \textbf{A--C} Per-axis coefficient distributions (box plots with overlaid points) for the mood (A), somatic (B), and suicidality (C) axes, with the in-sample Core Clinical corpus on the top row and the held-out Naturalistic corpus on the bottom; boxes show the median and interquartile range, points show individual excerpts, and $n$ is the number of excerpts per group. Brackets denote pairwise comparisons between annotation groups (***, $P<0.001$; **, $P<0.01$; *, $P<0.05$; ns, not significant). $P$-values are from Dunn's post-hoc test on pooled within-axis ranks, Benjamini--Hochberg-corrected jointly across all nine comparisons. \textbf{D} Median $\beta$ for each annotation group on each axis, with cells shaded and ranked by within-column standing (1st = highest median $\beta$); the bordered diagonal marks each group scoring highest on its own axis. The preserved within-corpus rank ordering is the evidence of generalization.}
\label{fig:coefficients}
\end{figure}

Having identified layer 21 as the operating layer, we next asked whether projecting text onto it yields per-symptom coefficients that agree with independent clinician annotation, and whether that agreement holds on a held-out set of naturalistic narratives. We answer these questions by projecting every passage from the three corpora onto the three layer-21 Symptom Vectors using the decorrelation procedure, yielding coefficients $\beta = (\beta_{\text{mood}}, \beta_{\text{somatic}}, \beta_{\text{suicidality}})$ per passage. As anticipated for textually comorbid symptom categories, the class-centroid basis was strongly collinear: the Gram matrix $G$ of the three Symptom Vectors had a condition number of 14{,}611.6, confirming $G$'s near-singularity, which would render a direct inverse numerically unstable and motivating the Moore--Penrose pseudoinverse correction. Table~\ref{tab:coefficients} reports the median together with Q1 and Q3 coefficients for each clinician-annotation group within each dataset; Fig.~\ref{fig:coefficients} shows the full per-axis distributions.

\begin{table}[!htbp]
\centering
\caption{Median projection coefficients by annotation group and symptom axis. Within each dataset, the annotation group matching a given symptom axis attains the highest median coefficient. This ordering holds from the in-sample Core Clinical corpus to the held-out Naturalistic corpus.}
\label{tab:coefficients}
\small
\resizebox{\textwidth}{!}{%
\begin{tabular}{llrrrr}
\toprule
Dataset & Annotation & $N$ & $\beta_{\text{mood}}$ median (Q1, Q3) & $\beta_{\text{somatic}}$ median (Q1, Q3) & $\beta_{\text{suicidality}}$ median (Q1, Q3) \\
\midrule
\multirow{3}{*}{Core Clinical $\cdot$ In-Sample}
 & Mood        & 19 & \textbf{+0.77 (+0.49, +1.14)} & +0.04 ($-$0.06, +0.20) & +0.10 ($-$0.00, +0.30) \\
 & Somatic     & 24 & $-$0.19 ($-$0.53, +0.21) & \textbf{+1.01 (+0.86, +1.27)} & +0.14 ($-$0.14, +0.38) \\
 & Suicidality & 8  & $-$0.05 ($-$0.27, +0.09) & +0.04 ($-$0.06, +0.08) & \textbf{+0.96 (+0.86, +1.34)} \\
\multirow{3}{*}{Naturalistic $\cdot$ Held-Out}
 & Mood        & 141 & \textbf{$-$0.08 ($-$0.33, +0.15)} & +0.22 (+0.11, +0.33) & +0.85 (+0.66, +1.08) \\
 & Somatic     & 141 & $-$0.64 ($-$1.04, $-$0.33) & \textbf{+0.74 (+0.61, +0.88)} & +0.91 (+0.58, +1.28) \\
 & Suicidality & 141 & $-$0.74 ($-$1.30, $-$0.28) & +0.34 (+0.23, +0.46) & \textbf{+1.40 (+0.95, +1.89)} \\
\bottomrule
\end{tabular}}
\end{table}

Within the Core Clinical corpus, the projection was internally consistent (Fig.~\ref{fig:coefficients}A--C): along each symptom axis the excerpts carrying the matching clinician annotation attained the highest coefficient. Mood-annotated excerpts projected most strongly onto the mood axis ($\beta_{\text{mood}}$ median $+0.77$), relative to somatic- and suicidality-annotated excerpts ($-0.19$ and $-0.05$); somatic-annotated excerpts led on the somatic axis ($+1.01$); and suicidality-annotated excerpts led on the suicidality axis ($+0.96$). As an in-sample consistency check, each annotation group scores highest on its own symptom axis relative to the other groups, indicating Symptom Vectors recover the categories from which they were constructed.

The same rank ordering was preserved in the held-out Naturalistic corpus (Fig.~\ref{fig:coefficients}A--C), the true test of generalization. On each axis, the matching-annotation group again attained the maximum coefficient: on the mood axis, mood-annotated excerpts scored highest ($\beta_{\text{mood}}$ median $-0.08$) above somatic- and suicidality-annotated excerpts ($-0.64$ and $-0.74$); on the somatic axis, somatic-annotated excerpts scored highest ($+0.74$) above mood- and suicidality-annotated excerpts ($+0.22$ and $+0.34$); and on the suicidality axis, suicidality-annotated excerpts scored highest ($+1.40$) above mood- and somatic-annotated excerpts ($+0.85$ and $+0.91$). Projecting held-out naturalistic narratives onto Symptom Vectors constructed from clinical instruments yields coefficients that agree with independent clinician annotation.

\begin{figure}[!htbp]
\begin{figpanel}[0.68]
  \includegraphics[width=\linewidth]{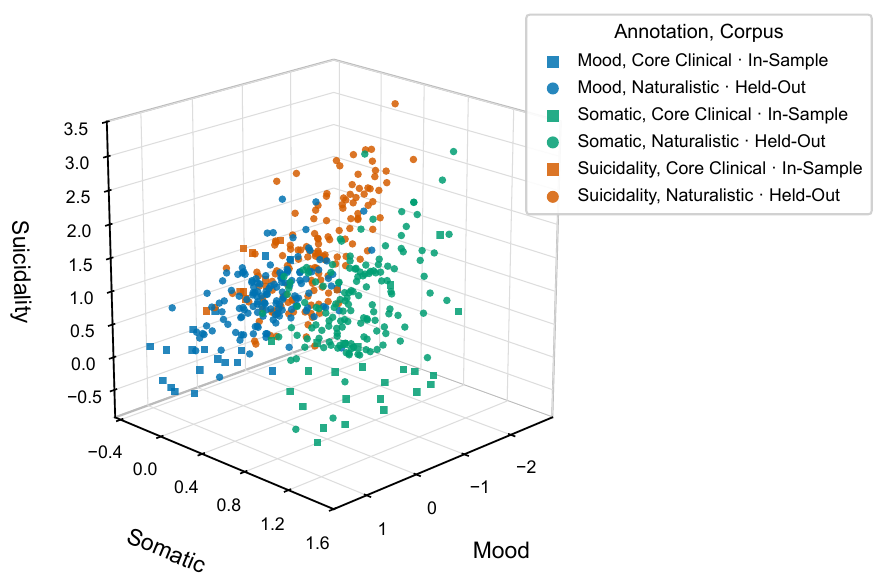}
\end{figpanel}%
\begin{figtext}
  \caption{\textbf{Passages sharing a clinician annotation occupy the same region of the symptom subspace.} Distribution of the three decorrelated projection coefficients $(\beta_{\text{mood}}, \beta_{\text{somatic}}, \beta_{\text{suicidality}})$ for all passages, plotted in the symptom subspace at layer 21. Each point is one passage, colored by clinician annotation (mood, somatic, suicidality) and shaped by corpus (squares, in-sample Core Clinical; circles, held-out Naturalistic).}
  \label{fig:subspace}
\end{figtext}
\end{figure}

The joint structure of these coefficients is visible in the three-dimensional projection (Fig.~\ref{fig:subspace}): clinical and naturalistic excerpts sharing an annotation co-localize within the symptom subspace.

The Gram-corrected projection recovered clinician annotations in-sample and preserved the same per-symptom rank ordering on held-out naturalistic text, demonstrating that Symptom Vectors distilled from clinical instruments generalize to naturalistic narratives.

\subsection{A single depression vector distinguishes depressive and non-depressive text}

\begin{figure}[!htbp]
\begin{figpanel}[0.68]
  \includegraphics[width=\linewidth,trim=0 20 16 12,clip]{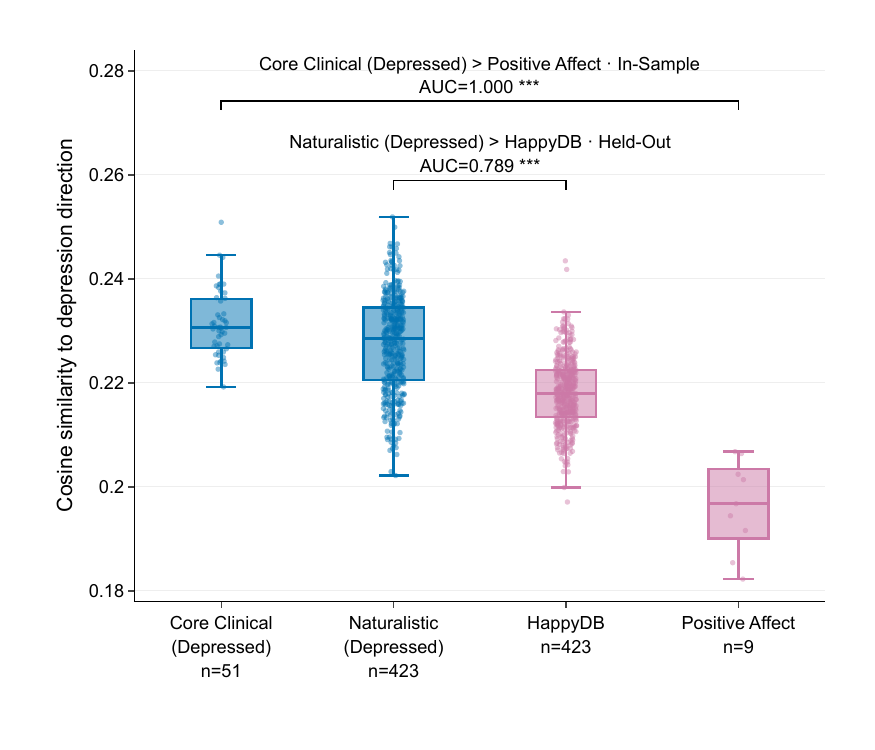}
\end{figpanel}%
\begin{figtext}
  \caption{\textbf{A consolidated layer-21 depression direction separates depressive from non-depressive text.} Cosine similarity between each excerpt's token-weighted centroid and the Depression Vector, by corpus. Core Clinical and Positive Affect are in-sample (both were used to construct the Depression Vector); Naturalistic and HappyDB are held out. Boxes show the median and interquartile range; points show individual excerpts. Brackets report AUC for the in-sample and held-out contrasts; ***, $P<0.001$, one-sided Mann--Whitney $U$ test.}
  \label{fig:depvector}
\end{figtext}
\end{figure}

Having established that symptom categories are separable and projectable at layer 21 (Figs.~\ref{fig:coefficients} and \ref{fig:subspace}), we finally asked whether the same representational space supports the distinction between depressive and non-depressive text. We scored every excerpt by the cosine similarity between its token-weighted centroid and the Depression Vector (Fig.~\ref{fig:depvector}).

Core Clinical excerpts ($n=51$) and Positive Affect excerpts ($n=9$) were completely separated by the Depression Score (AUC~$=$~1.000, one-sided Mann--Whitney $U$, $P=1.06\times10^{-6}$). Since both corpora participated in constructing the axis, this result serves only as an in-sample consistency check, not as evidence of generalization.

Neither the held-out Naturalistic nor the HappyDB corpus contributed to the construction of the Depression Vector, so the difference between them tests generalization of the valence axis to unseen text. Naturalistic depressive excerpts ($n=423$) received higher Depression Scores than HappyDB happy-moment descriptions ($n=423$), with an AUC of 0.789 (one-sided Mann--Whitney $U$, $P=2.51\times10^{-48}$): a randomly drawn depressive narrative outscores a randomly drawn happy-moment description roughly four times out of five.

Notably, the two HappyDB outliers with visibly elevated Depression Scores are neutral in emotional valence. The highest-scoring entry reads, ``Their sore throat cleared up''; the second highest, ``They went shopping.''

Two features of the score distributions are worth noting. First, the four corpora ordered as expected along the axis, with median cosine similarity decreasing monotonically from Core Clinical (0.231) to Naturalistic (0.229) to HappyDB (0.218) to Positive Affect (0.197), consistent with the Depression Vector capturing a graded valence signal rather than a corpus-identity artifact. Second, the held-out distributions overlap substantially: many HappyDB excerpts score within the Naturalistic range, and the absolute cosine values span a narrow band (0.18--0.25). The Depression Score therefore separates depressive from non-depressive text at the population level but is not, on its own, a reliable per-excerpt classifier; its role here is to establish that a valence gate is recoverable from the same layer-21 geometry used for symptom projection.

\section{Discussion}

One challenge in treating depression is its clinical heterogeneity: patients sharing a diagnosis exhibit widely divergent symptom profiles. Yet, standard questionnaires often collapse this complexity into a single severity score\cite{Fried2015,Chevance2020}. Here, we show that when a pre-trained large language model (Gemma-3-27B-PT) processes patient narratives, distinct depression symptoms are recoverable directly from its residual stream. Three key findings support this conclusion. First, the model's internal representations of different symptom categories (mood/emotional/cognitive, somatic, and suicidality) are geometrically separable. This separability is most prominent at layer 21 across three raw distance metrics (Centroid Euclidean, Energy, and Earth Mover's; pseudo-$F = 6.91$, $6.49$, and $2.83$, respectively), driven by distinct group centroids rather than within-group dispersion (all PERMDISP $P \ge 0.43$). Second, projecting new patient narratives onto these layer-21 Symptom Vectors with a Gram-pseudoinverse correction yields decorrelated coefficients that preserve clinician-aligned rank ordering on held-out naturalistic narratives. Finally, because these Symptom Vectors are built entirely from residual stream activations prompted by depressive text, their projection coefficients are only meaningful for depressive speech. To address this caveat, we introduce a ``Depression Vector,'' also derived from layer-21 activations, that acts as a preliminary valence gate to first identify whether the input text is depressive. Ultimately, this two-stage pipeline extracts decorrelated, clinician-aligned symptom profiles of patient speech directly from LLM activations.

These findings extend a growing body of work showing that semantic categories are encoded as approximately linear directions in LLM residual streams\cite{Park2025}. Among these categories is sentiment, which is localized to specific layers\cite{Hollinsworth2024,DiPalma2025}. While prior work has characterized coarse valence, we map an instrument-derived symptom taxonomy (DSM-5, ICD-10, HAM-D, MADRS, PHQ-9, and PROCEED) onto a model's internal representations and recover symptom signals along three axes. Our approach substantially differs from prior systems that rely on generated output for broad diagnostic labels\cite{Yang2024,Xu2024,Lamichhane2023} or require task-specific fine-tuning for symptom-level analysis\cite{Weber2025}. Instead, we extract the signal directly from the internal activations of an unmodified pre-trained model.

The Gram-pseudoinverse decorrelation of the three strongly collinear symptom axes (Gram matrix condition number $\approx 1.5\times10^4$) turns overlapping clinical categories into a stable set of measurement axes, so that text excerpts from different sources can be scored and compared on the same scale. The Depression Vector, by contrast, follows the semantic-projection framework of Grand et al.\cite{Grand2022} closely: because depressive and positive affect are opposed in emotional valence, a contrastive axis between them is well defined, and a simple cosine projection onto it generalizes to unseen text.

Interpretability is a major barrier to clinical trust and the translation of model-based tools into care. Our pipeline addresses this opacity by directly extracting symptom representations from residual stream activations. The practical value of this approach for digital medicine is that the signal is immediately available within the activation space, eliminating the need to inspect generated output. In practice, this pipeline should first apply a depressive valence gate to prevent assigning symptom scores to non-depressive speech by using the Depression Vector. The Symptom Vectors we found could also help fine-tune models to respond more safely to vulnerable users. Unlike prior steering methods, which typically rely on a single general-purpose direction\cite{Panickssery2024,Zou2023}, they offer multiple directions grounded in clinical categories. Finally, these decorrelated coefficients would allow us to link LLM symptom signals and neurobiological biomarkers. Using LLMs to continuously extract per-symptom scores from naturalistic speech creates a quantitative signal that can be correlated with fMRI, resting EEG, or TMS-evoked potentials\cite{Parmigiani2023}. This methodology builds on established approaches that successfully mapped continuous behavioral markers to neural activity in naturalistic settings, such as decoding facial expressions for pain\cite{Huang2025} and mood\cite{Kakusa2025}. By applying this approach to language, we can test whether the distinct mood and somatic dimensions extracted by the LLM correspond to dissociable neural signatures. If successful, these language models may provide scalable, low-cost proxies for complex brain imaging, advancing the symptom-specific targets of precision psychiatry\cite{Cline2026}.

Several limitations must be addressed prior to clinical application. First, we collapsed the nine DSM-5 symptoms into three groups for statistical power, which limits the symptom specificity of our analysis. Second, all analyses derive from a single model; generalization across architectures and scales is untested. Third, preprocessing is a potential confound: every passage was rewritten into third-person prose by another large language model. Although the edits were minimal, reviewed by psychiatrist authors, and designed to preserve clinical content, this step may introduce bias. Fourth, and most importantly, agreement with clinician annotation is not equivalent to clinical validity: nothing reported here has been validated against diagnostic outcomes, treatment response, or longitudinal trajectories, and the method should not be construed as a clinical instrument. The prospect of an activation-level risk detector raises questions of consent, false-positive burden, and surveillance that any deployment would need to resolve through appropriate governance.

Future research should address these limitations by granularizing Symptom Vectors using larger clinical corpora, replicating localization results across diverse model families and scales, calibrating projections into an absolute and longitudinally stable severity scale suitable for patient monitoring, and empirically testing the proposed electrophysiological links and risk-detection applications. With these steps, the clinician-aligned Symptom Vectors identified here provide a mechanistic and interpretable foundation for symptom analysis of patient language, and a generalizable template for extracting other clinical concepts from the internal representations of large language models.

\section*{Data availability}
The third-person rewritten Core Clinical corpus (excluding copyright-protected source text) and the Positive Affect corpus (public-domain sources) are released at the project repository, together with the layer-21 Symptom Vectors and the Gram-corrected projection pipeline. The following corpora are withheld: raw clinical instrument text is not redistributed in accordance with source copyright, and the ReDSM5 corpus is not redistributed under its license terms; \textit{Darkness Visible} and the \textit{Handbook of Depression} excerpts are also copyright-protected and not redistributable. Citations are provided in the Methods to allow independent assembly of an equivalent corpus.

\section*{Code availability}
\begin{sloppypar}
All analyses were performed in Python. Code is available at \url{https://github.com/PrecisionNeuroLab/Symptom-Vectors-for-Depression}.
\end{sloppypar}

\section*{Ethics declarations}
This study involved the secondary analysis of publicly available, de-identified datasets. Data sources included HappyDB, a corpus of crowdsourced text responses, and ReDSM5, a dataset of public social media posts. Because the research relied exclusively on pre-existing, de-identified, and publicly accessible data, the study was deemed exempt from Institutional Review Board (IRB) review.

\section*{GenAI usage disclosure}
Anthropic Claude Opus 4 has been used to pre-process text samples. Figure~\ref{fig:pipeline} is created with Claude Design.

Generative AI tools were employed for coding assistance. All AI-written code has been reviewed, tested, and verified by the authors.

During the preparation of this manuscript, generative AI tools were employed for editing purposes, including proofreading, grammar correction, vocabulary improvement, and overall language polishing.

\section*{Acknowledgements}
We thank P.\,J.\ Hansel of LeafEye for providing the Apple Silicon hardware on which the experiments were run, and Connor Stone (University of Toronto) for collaboration on the \texttt{pted} energy-distance package. We also thank P.\,J.\ Hansel and Ethan Andrew Solomon (Stanford University) for providing valuable feedback on the manuscript.

\section*{Author contributions}
All authors contributed to writing the manuscript. F.Z.\ wrote the analysis code, preprocessed all three datasets, conducted all experiments, and wrote the manuscript. A.S.\ conceived the study and participated in all experiment designs. A.C.\ and C.W.\ identified symptom categories using clinical expertise and reviewed and edited the manuscript. R.G.\ helped with fact checking, reviewing, and editing parts of the manuscript. C.J.K.\ designed part of the experiments, data analysis and presentation, and supervised the study.

\section*{Competing interests}
F.Z.\ and R.G.\ hold stock in Alphabet Inc., the parent company of Google, which developed the Gemma model evaluated in this study. A.S.\ holds stock in Orchard Neuro. C.J.K.\ is a consultant for Salma Health, Flow Neuroscience, Kyron Medical, and Constellation Systems, and holds stock in Orchard Neuro.

\newpage
\appendix
\section*{Supplementary Material}

\subsection*{Core Clinical Symptom Descriptions}

Natural-language descriptions of depression symptoms, organized by clinical assessment instrument and grouped into three symptom dimensions (mood, somatic, and suicidality). Each entry corresponds to an item or item cluster from the source instrument.

\subsubsection*{DSM-5}
\symgroup{Mood}
This person experiences a depressed mood, feeling sad, empty, or hopeless, and may appear tearful to others. They have lost interest or pleasure in activities they used to enjoy. They struggle with feelings of worthlessness or carry excessive or inappropriate guilt that goes beyond normal self-criticism. They also have difficulty thinking clearly, concentrating, or making decisions.

\symgroup{Somatic}
They experience noticeable changes in weight---either losing or gaining---without intentionally trying to diet, along with shifts in their appetite. Their sleep is disrupted, with difficulty falling or staying asleep, or alternatively, sleeping far more than usual. Others can observe changes in their physical movements and behavior: they may appear restless and agitated, or conversely, slowed down in their speech and actions. They feel persistently tired and lack the energy they once had.

\symgroup{Suicidality}
The person frequently thinks about death, beyond simply fearing dying. They may have recurring thoughts of suicide without a specific plan, or they may have attempted suicide or developed a specific plan for committing suicide.

\subsubsection*{ICD-10}
\symgroup{Mood}
They experience a persistently low mood throughout the day, which others may also observe. They have lost interest or pleasure in activities they would normally enjoy. They struggle with feelings of worthlessness and carry excessive or inappropriate guilt. Their ability to think clearly or concentrate is diminished, and they often feel indecisive. Their self-confidence and self-esteem are reduced.

\symgroup{Somatic}
They experience changes in their appetite---either eating much less or much more than usual---which leads to noticeable weight loss or gain. Their sleep is disrupted; they may struggle to fall or stay asleep, or conversely find themselves sleeping far more than normal. Even minimal effort leaves them feeling tired and drained, as though their energy has been depleted. Their physical movements and behavior have also changed---they may appear restless and agitated, or alternatively seem slowed down in their speech and actions.

\symgroup{Suicidality}
They experience recurring thoughts about death or have engaged in suicidal behavior.

\subsubsection*{HAM-D (Hamilton Depression Rating Scale)}
\symgroup{Mood}
\paragraph{Item 1. Depressed Mood} They have a gloomy attitude and feel pessimistic about the future. They experience persistent feelings of sadness and may have a tendency to weep.
\paragraph{Item 2. Feelings of Guilt} They engage in self-reproach and feel they have let people down. They experience ideas of guilt, and may believe their present illness is a punishment. In some cases, they may have delusions or hallucinations related to guilt.
\paragraph{Item 3. Work and Interests} They experience feelings of incapacity, listlessness, indecision, and vacillation. They have lost interest in hobbies and show decreased social activities. Their productivity has decreased, and they may be unable to work due to their present illness.
\paragraph{Item 4. Anxiety (Psychic)} They experience tension and irritability. They worry about minor matters and have an apprehensive attitude. They may experience fears.

\symgroup{Somatic}
\paragraph{Item 1. Initial Insomnia} The person has difficulty falling asleep at the beginning of the night.
\paragraph{Item 2. Middle Insomnia} The person feels restless and disturbed during the night, often waking up during sleeping hours.
\paragraph{Item 3. Delayed Insomnia} The person wakes up in the early hours of the morning and is unable to fall asleep again.
\paragraph{Item 4. Retardation} The person experiences slowness in their thought processes, speech, and physical activity. They may appear apathetic or, in more pronounced cases, enter a state of stupor.
\paragraph{Item 5. Agitation} The person displays restlessness that is associated with feelings of anxiety.
\paragraph{Item 6. Somatic Anxiety} The person experiences physical symptoms related to anxiety, which may include gastrointestinal issues such as indigestion, cardiovascular symptoms like palpitations, headaches, respiratory difficulties, or genito-urinary complaints.
\paragraph{Item 7. Gastrointestinal Somatic Symptoms} The person experiences a loss of appetite, a heavy feeling in the abdomen, or constipation.
\paragraph{Item 8. General Somatic Symptoms} The person feels heaviness in their limbs, back, or head, and may experience diffuse backaches, a loss of energy, or easy fatiguability.
\paragraph{Item 9. Genital Symptoms} The person experiences a loss of libido or, in the case of women, menstrual disturbances.
\paragraph{Item 10. Hypochondriasis} The person shows excessive focus on their body and health, ranging from bodily self-absorption to preoccupation with health concerns, a tendency to complain about physical ailments, or fixed false beliefs about having a serious illness.
\paragraph{Item 11. Weight Loss} The person has experienced a noticeable decrease in body weight.

\symgroup{Suicidality}
The person may experience thoughts related to not wanting to live. This can include feeling that life is not worth living, wishing they were dead, having suicidal ideas or making suicidal gestures, or attempting suicide.

\subsubsection*{MADRS (Montgomery--\r{A}sberg Depression Rating Scale)}
\symgroup{Mood}
\paragraph{Item 1. Apparent Sadness} They appear despondent, gloomy, or in despair---beyond ordinary low spirits. This is reflected in their speech, facial expression, and posture. They may look dispirited or unhappy, and at times struggle to brighten up even when circumstances might warrant it.
\paragraph{Item 2. Reported Sadness} They describe feeling depressed, low in spirits, or despondent, regardless of whether this is visible in their appearance. They may report feeling beyond help or without hope. Their mood may be influenced by external circumstances to varying degrees, or it may feel continuous and unvarying.
\paragraph{Item 3. Inner Tension} They experience feelings of ill-defined discomfort, edginess, or inner turmoil. This mental tension can escalate to feelings of panic, dread, or anguish. They may seek reassurance and find it difficult to master these feelings on their own.
\paragraph{Item 4. Concentration Difficulties} They have difficulty collecting their thoughts or sustaining concentration. This may interfere with their ability to read, hold a conversation, or complete tasks that require focused attention.
\paragraph{Item 5. Inability to Feel} They experience reduced interest in their surroundings or in activities that would normally bring them pleasure. Their ability to react with appropriate emotion to circumstances or people feels diminished. They may feel emotionally numb or disconnected from friends and loved ones, struggling to feel anger, grief, or joy.
\paragraph{Item 6. Pessimistic Thoughts} They experience thoughts of guilt, inferiority, self-reproach, or remorse. They may engage in self-accusations or feel increasingly pessimistic about the future, dwelling on ideas of failure, sin, or ruin.

\symgroup{Somatic}
\paragraph{Item 1. Reduced Sleep} They experience a reduction in the duration or depth of their sleep compared to their usual pattern when feeling well. This may manifest as difficulty falling asleep, lighter or more fitful sleep, sleep that is broken or shortened by several hours, or sleeping only a few hours per night.
\paragraph{Item 2. Reduced Appetite} They experience a diminished appetite compared to when they are well. This presents as a decreased desire for food or a need to force themselves to eat. Food may seem tasteless, and they may require encouragement or persuasion from others to eat at all.
\paragraph{Item 3. Lassitude} They have difficulty getting started with activities and experience slowness in initiating and performing everyday tasks. Routine activities require considerable effort to begin and carry out, and in more pronounced cases, they may feel unable to do anything without assistance from others.

\symgroup{Suicidality}
\paragraph{Item 1} This person has expressed feelings that life is not worth living and has indicated that a natural death would be welcome. They may feel weary of life, with fleeting or recurring thoughts of suicide.
\paragraph{Item 2} At times, they may believe they would be better off dead, viewing suicide as a possible solution, though not necessarily with specific plans or intention.
\paragraph{Item 3} They experience suicidal thoughts and may have made preparations for suicide. In some cases, they may have explicit plans for suicide or may be actively preparing for it.

\subsubsection*{PHQ-9 (Patient Health Questionnaire-9)}
\symgroup{Mood}
\paragraph{Item 1} They have little interest or pleasure in doing things.
\paragraph{Item 2} They feel down, depressed, or hopeless.
\paragraph{Item 3} They feel bad about themselves, or feel like a failure, or feel they have let themselves or their family down.
\paragraph{Item 4} They have trouble concentrating on things, such as reading the newspaper or watching television.

\symgroup{Somatic}
\paragraph{Item 1} They have trouble falling asleep, staying asleep, or sleep too much.
\paragraph{Item 2} They feel tired or have little energy.
\paragraph{Item 3} They have a poor appetite or overeat.
\paragraph{Item 4} They move or speak so slowly that other people could have noticed, or the opposite---they are so fidgety or restless that they have been moving around a lot more than usual.

\symgroup{Suicidality}
They have thoughts that they would be better off dead or of hurting themselves in some way.

\subsubsection*{PROCEED (Participative Research on Outcomes' and Core Expectations' Elicitation for Depression)}
\symgroup{Mood}
\paragraph{Item 1. Perception of Self} This person often feels that others are against them or out to get them. They experience a profound sense of isolation, as though they are fundamentally alone in the world. They carry a belief that they are a burden to those around them and feel that others do not truly understand them. They struggle with confidence in their own abilities and hold a diminished view of their own worth. At times, they feel disconnected from themselves or their surroundings, and they may have difficulty recognizing themselves---whether in the mirror or in terms of who they feel they are as a person.
\paragraph{Item 2. Cognitive Symptoms} This person finds it difficult to feel motivated or to maintain interest in activities. They believe their condition is untreatable and hold a generally pessimistic view of situations. Their thinking tends to be distorted, often interpreting events in negatively skewed ways. They struggle to make decisions and have difficulty with planning, organizing, and following through on tasks. Adapting to new information or shifting between tasks is challenging for them. They find it hard to solve problems effectively and experience lapses in memory. Maintaining focus and concentration is difficult, as is managing their time. Their perception of time may feel altered---either dragging or slipping away. They have trouble envisioning a positive future and tend to dwell on negative thoughts repeatedly. Their thinking often feels foggy or unclear. They may have reduced capacity to understand or share in others' feelings and show diminished interest in social connection.
\paragraph{Item 3. Mood and Emotional Symptoms} This person experiences significant psychological pain and a pervasive sense that things will not improve. They feel powerless to change their circumstances and see themselves as having little value. A general sense of feeling bad pervades their experience, often accompanied by guilt. Sadness is a frequent companion, along with feelings of anxiety and a fear of failing at things they attempt. They often feel restless and on edge, becoming easily irritated and frustrated. They have difficulty managing their emotions and controlling their impulses. Their mood may not shift appropriately in response to positive events. They have lost their sense of humor and find little or no pleasure in activities they once enjoyed. They often feel empty inside, and their emotional responses in general feel muted or dulled.

\symgroup{Somatic}
\paragraph{Item 1. Physical Symptoms} They experience pain or other physical symptoms. Their appetite or weight has changed. They have difficulty with sexual function.
\paragraph{Item 2. Sleep} They sleep excessively. Their sleep is disturbed or restless. They have difficulty falling or staying asleep.
\paragraph{Item 3. Psychomotor and Expression} They cry. They speak very little or not at all. Their facial expressions appear flat or diminished. Their movements and reactions are slowed.
\paragraph{Item 4. Energy} They feel that everything requires effort. They feel physically weak. They have low energy. They feel fatigued or tired.

\symgroup{Suicidality}
\paragraph{Autoaggression symptoms} The individual experiences thoughts related to their desire to continue living. They report suicidal ideation and engage in self-harming behaviors. They have attempted suicide.

\end{document}